\documentclass[runningheads]{llncs}
\usepackage{graphicx}
\usepackage{hyperref}

\usepackage{amsmath,amssymb,amsfonts}
\usepackage{algorithmic}
\usepackage{textcomp}
\usepackage{xcolor}
\def\BibTeX{{\rm B\kern-.05em{\sc i\kern-.025em b}\kern-.08em
    T\kern-.1667em\lower.7ex\hbox{E}\kern-.125emX}}
\newcommand\fnurl[2]{\href{#2}{#1}\footnote{\url{#2}}}
\newcommand{\etal} {\textit{et~al.}}
\usepackage{url}
\usepackage{booktabs}
\usepackage{array}
\usepackage{stfloats}

\begin{document}
\title{Text line extraction using fully convolutional network and energy minimization}
%
%
\author{Berat Kurar Barakat\inst{1}\orcidID{0000-0002-7240-7286} \and
Ahmad Droby\inst{1} \and
Reem Alaasam\inst{1} \and
Boraq Madi\inst{1} \and
Irina Rabaev\inst{2} \and
Jihad El-Sana\inst{1}}
\authorrunning{F. Author et al.}
%
\institute{Ben-Gurion University of the Negev\\
\email{\{berat,drobya,rym,borak\}@post.bgu.ac.il} \and
Shamoon College of Engineering\\
\email{\{irinar\}@ac.sce.ac.il}}
\maketitle              
\begin{abstract}
Text lines are important parts of handwritten document images and easier to analyze by further applications. Despite recent progress in text line detection, text line extraction from a handwritten document remains an unsolved task. This paper proposes to use a fully convolutional network for text line detection and energy minimization for text line extraction. Detected text lines are represented by blob lines that strike through the text lines. These blob lines assist an energy function for text line extraction. The detection stage can locate arbitrarily oriented text lines. Furthermore, the extraction stage is capable of finding out the pixels of text lines with various heights and interline proximity independent of their orientations. Besides, it can finely split the touching and overlapping text lines without an orientation assumption. We evaluate the proposed method on VML-AHTE, VML-MOC, and Diva-HisDB datasets. The VML-AHTE dataset contains overlapping, touching and close text lines with rich diacritics. The VML-MOC dataset is very challenging by its multiply oriented and skewed text lines. The Diva-HisDB dataset exhibits distinct text line heights and touching text lines. The results demonstrate the effectiveness of the method despite various types of challenges, yet using the same parameters in all the experiments.

\keywords{Historical documents analysis \and Text line segmentation\and Text line detection\and Text line extraction\and Handwritten document}
\end{abstract}

\section{Introduction}
\label{introduction}

Segmentation in computer vision is the task of dividing an image into parts that are easier to analyse. Text lines of a handwritten document image are widely used for word segmentation, text recognition and spotting, manuscripts alignment and writer recognition. Text lines need to be provided to these applications either by their locations or by complete set of their pixels. The task of identifying the location of each text line is called detection, whereas the task of determining the pixels of each text line is called extraction. Much research in the recent years has focused on text line detection~\cite{renton2017handwritten,barakat2018text,gruning2019two}. However, detection defines the text lines loosely by baselines or main body blobs. On the other hand, extraction is a harder task which defines text lines precisely by pixel labels or bounding polygons. 

The challenges in text line extraction arise due to variations in text line heights and orientations, presence of overlapping and touching text lines, and diacritical marks within close interline proximity. It has been generally demonstrated that deep learning based methods are effective at detecting text lines with various orientations \cite{vo2016dense,oliveira2018dhsegment,renton2018fully,gruning2019two}. However, only few of the recent researches \cite{campos2018text,vo2016dense} have addressed the problem of extraction given the detection, yet with the assumption of horizontal text lines.

\begin{figure}[h]
\centering
\includegraphics[width=10.5cm]{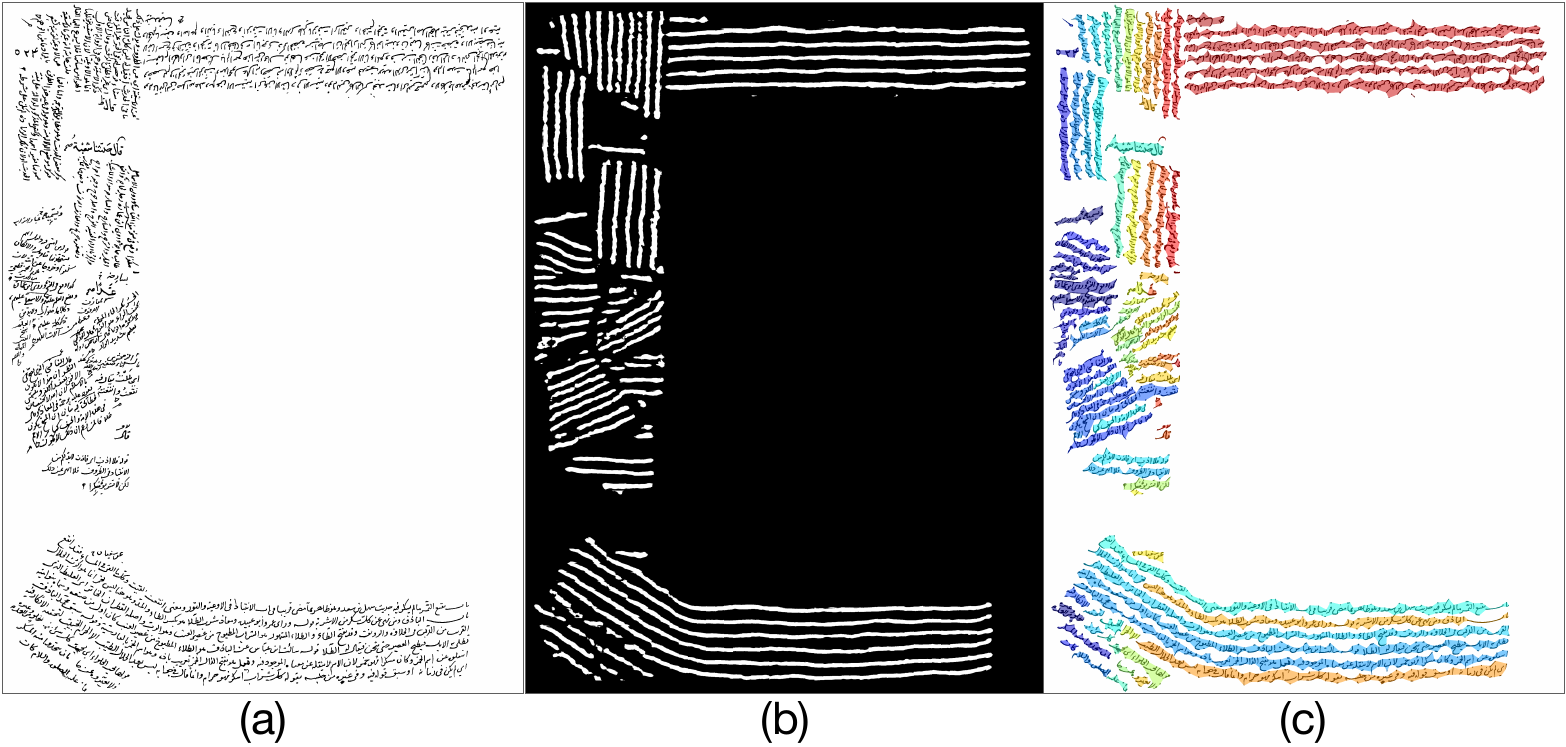}
\caption{Given a handwritten document image (a), FCN learns to detect blob lines that strike through text lines (b). EM with the assistance of detected blob lines extracts the pixel labels of text lines which are in turn enclosed by bounding polygons (c).}
\label{phases}		
\end{figure}

This paper proposes a text line extraction method (FCN+EM) which uses Fully Convolutional Network (FCN) to detect text lines in the form of blob lines (\figurename~\ref{phases}(b)), followed by an Energy Minimization (EM) function assisted by these blob lines to extract the text lines (\figurename~\ref{phases}(c)). FCN is capable of handling curved and arbitrarily oriented text lines. However, extraction is problematic due to the Sayre's paradox \cite{sayre1973machine} which states that exact boundaries of handwritten text can be defined only after its recognition and handwritten text can be recognized only after extraction of its boundaries. Nevertheless, humans are good at understanding boundaries of text lines written in a language they do not know. Therefore, we consider EM framework to formulate the text line extraction in compliance with the human visual perception, with the aid of the Gestalt proximity principle for grouping \cite{koffka2013principles}. The proposed EM formulation for text line extraction is free of an orientation assumption and can be used with touching and overlapping text lines with disjoint strokes and close interline proximity (\figurename~\ref{phases}(a)).

The proposed extraction method (FCN+EM) is evaluated on Visual Media Lab Arabic Handwritten Text line Extraction (VML-AHTE) dataset, Multiply Oriented and Curved (VML-MOC) dataset \cite{barakat2019vml}, and DIVA Historical Manuscript Database (DIVA-HisDB) \cite{simistira2017icdar2017}. VML-AHTE dataset is characterized by touching and overlapping text lines with close proximity and rich diacritics. VML-MOC dataset contains arbitrarily oriented and curved text lines. DIVA-HisDB dataset exhibit varying text line heights and touching text lines.

The rest of the paper is organized as follows. Related work is discussed in Section \ref{relatedwork}, and the datasets are described in Section \ref{datasets}. Later, the method is presented in Section \ref{method}. The experimental evaluation and the results are then provided in Section \ref{experiments}. Finally, Section \ref{conclusion} draws conclusions and outlines future work.

%
%
%
%
%
%
%
\section{Related work}
\label{relatedwork}

A text line is a set of image elements, such as pixels or connected components. Text line components in a document image can be represented using basic geometric primitives such as points, lines, polylines, polygons or blobs. Text line representation is given as an input to other document image processing algorithms, and, therefore, important to be complete and correct.


There are two main approaches to represent text lines: text line detection and text line extraction. Text line detection detects the lines, polylines or blobs that represent the locations of spatially aligned set of text line elements. Detected line or polyline is called a baseline~\cite{renton2017handwritten,gruning2019two} if it joins the lower part of the character main bodies, and a separator path~\cite{saabni2014text,campos2018text} if it follows the space between two consecutive text lines. Detected blobs~\cite{barakat2018text} that cover the character main bodies in a text line are called text line blobs. 

Text line extraction determines the constituting pixels or the polygons around the spatially aligned text line elements. Pixel labeling assigns the same label to all the pixels of a text line~\cite{saabni2014text,cohen2014using,vo2016dense}. Bounding polygon is used to enclose all the elements of a text line together with its neighbourhood background pixels~\cite{fischer2014combined,gruuening2017robust}. Most of the extraction methods assume horizontally parallel text lines with constant heights, whereas some methods~\cite{barakat2019vml,aldavert2018manuscript} are more generic.

Recent deep learning methods estimate $x$-height of text lines using FCN and apply Line Adjacency Graphs (LAG) to post-process FCN output to split touching lines~\cite{moysset2015paragraph,moysset2016learning}. Renton~\etal~\cite{renton2017handwritten,renton2018fully} also use FCN to predict $x$-height of text lines. Kurar~\etal~\cite{barakat2018text} applied FCN for challenging manuscript images with multi-skewed, multi-directed and curved handwritten text lines. However these methods either do only text line detection or their extraction phase is not appropriate for unstructured text lines because their assumption of horizontal and constant height text lines. The proposed method assumes the both, detection and extraction phases to be for complex layout.

ICDAR 2009~\cite{gatos2011icdar2009} and ICDAR 2013~\cite{stamatopoulos2013icdar} datasets are commonly used for evaluating text line extraction methods and ICDAR 2017~\cite{diem2017cbad} dataset is used for evaluating text line detection methods. DIVA-HisDB dataset~\cite{simistira2017icdar2017} is used for both types of evaluations: detection and extraction. Therefore, we select to use DIVA-HisDB~\cite{simistira2017icdar2017} as it provides ground truth for detection and extraction. However, this dataset is not enough representative of all the segmentation problems to evaluate a generic method. Hence, we also evaluated the proposed method on publicly available VML-MOC~\cite{barakat2019vml} dataset that contains multiply oriented and curved text lines with heterogeneous heights, and on VML-AHTE dataset that contains crowded diacritics.

\section{Datasets}
\label{datasets}
We evaluated the proposed method on three publicly available handwritten datasets. We suppose that these datasets demonstrate the generality of our method. As VML-AHTE dataset contains lines with crowded diacritics, VML-MOC dataset contains multiply oriented and curved lines, and Diva-HisDB dataset contains consecutively touching multiple lines. In this section we present these datasets.

\subsection{VML-AHTE}
\begin{figure}[h]
\centering
\includegraphics[width=7cm]{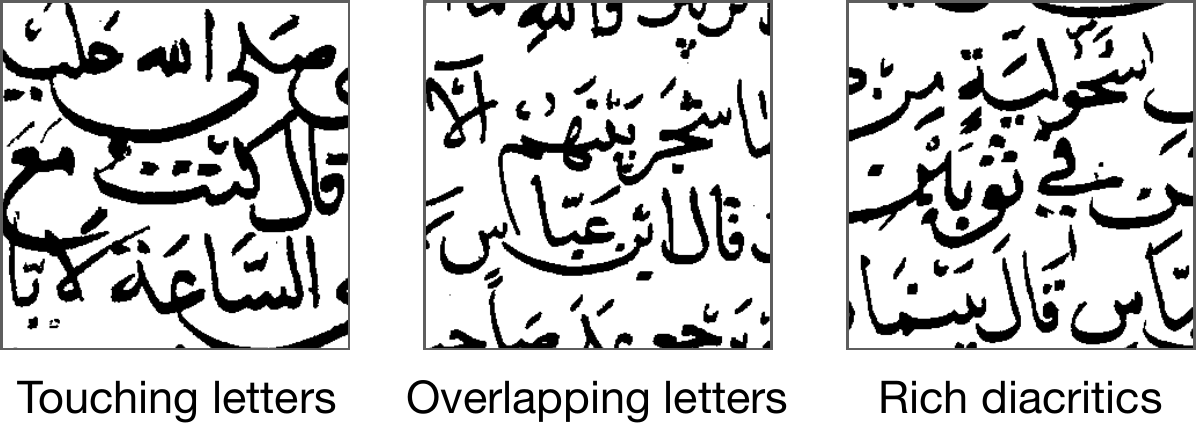}
\caption{Some samples of challenges in VML-AHTE dataset.}
\label{ahte_dataset}		
\end{figure}

VML-AHTE dataset is a collection of 30 binary document images selected from several manuscripts (Fig.~\ref{ahte_dataset}). It is a newly published dataset and available online for \fnurl{downloading}{https://www.cs.bgu.ac.il/~berat/data/ahte_dataset}. The dataset is split into 20 train pages and 10 test pages. Its ground truth is provided in three formats: bounding polygons in PAGE xml~\cite{pletschacher2010page} format, color pixel labels and DIVA pixel labels~\cite{simistira2017icdar2017}.

\subsection{Diva-HisDB}
\begin{figure}[h]
\centering
\includegraphics[width=7cm]{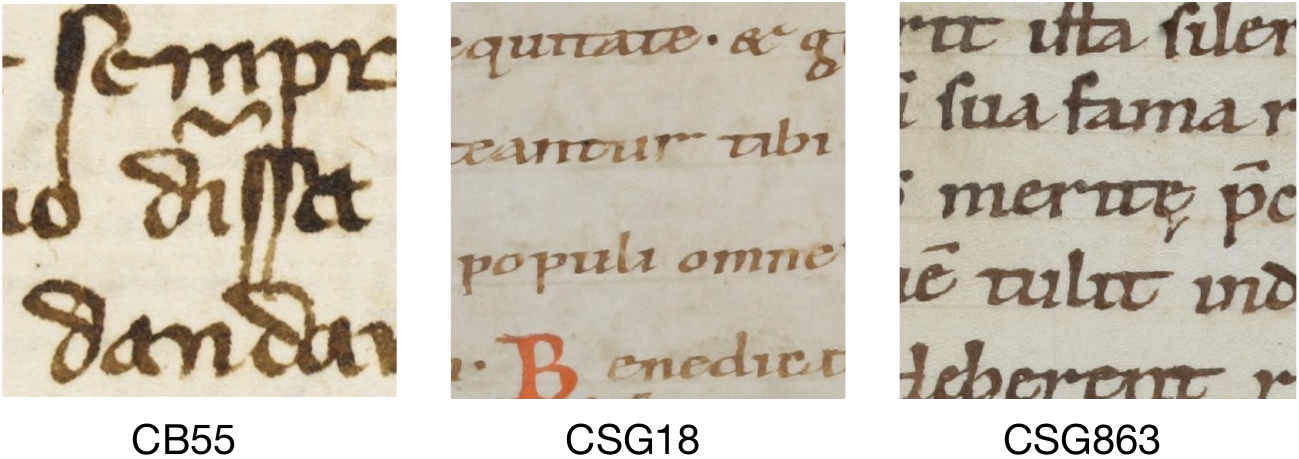}
\caption{Diva-HisDB dataset contains 3 manuscripts: CB55, CSG18 and CSG863. Notice the touching characters among multiple consecutive text lines in CB55.} 
\label{diva_dataset}
\end{figure}

DIVA-HisDB dataset \cite{simistira2017icdar2017} contains 150 pages from 3 medieval manuscripts: CB55, CSG18 and CSG863 (Fig.~\ref{diva_dataset}). Each book has 20 train pages and 10 test pages. Among them, CB55 is characterized by a vast number of touching characters. Ground truth is provided in three formats: baselines and bounding polygons in PAGE xml~\cite{pletschacher2010page} format and DIVA pixel labels~\cite{simistira2017icdar2017}.

\subsection{VML-MOC}
\begin{figure}
\centering
\includegraphics[width=10cm]{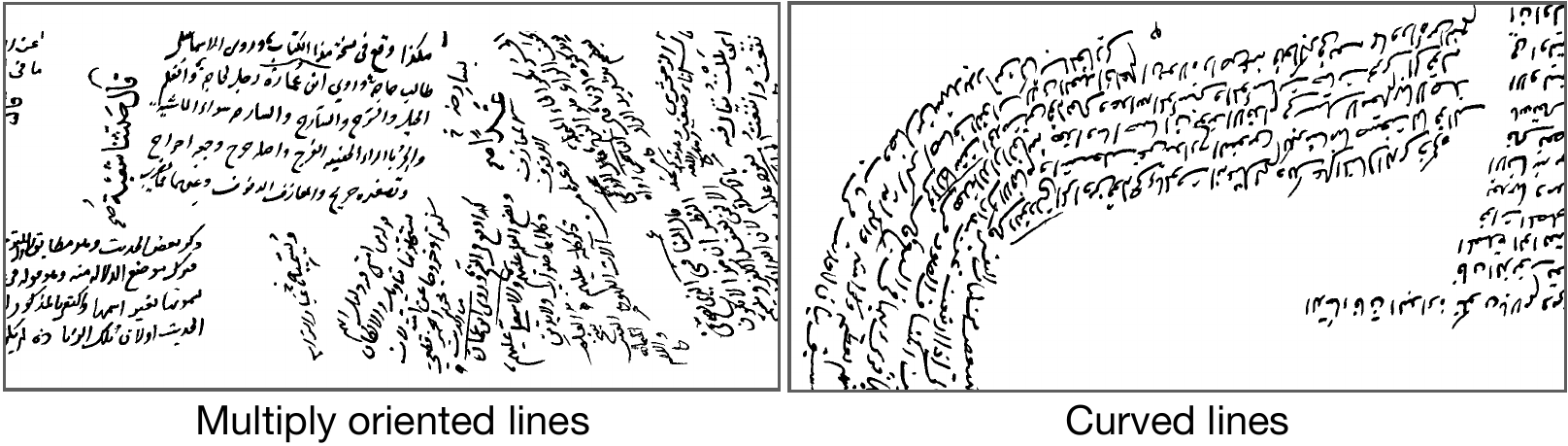}
\caption{VML-MOC dataset purely contains binarized side notes with arbitrary orientations and curvy forms.}
\label{moc_dataset}		
\end{figure}

VML-MOC dataset \cite{barakat2019vml} is a multiply oriented and curved handwritten text lines dataset that is publicly \fnurl{available}{https://www.cs.bgu.ac.il/~berat/data/moc_dataset}. These text lines are side notes added by various scholars over the years on the page  margins, in arbitrary orientations and curvy forms due to space constraints (Fig.~\ref{moc_dataset}). The dataset contains 30 binary document images and divided into 20 train pages and 10 test pages. The ground truth is provided in three formats: bounding polygons in PAGE xml~\cite{pletschacher2010page} format, color pixel labels and DIVA pixel labels~\cite{simistira2017icdar2017}.

\section{Method}
\label{method}

We present a method (FCN+EM) for text line detection together with extraction, and show its effectiveness on handwritten document images. In the first phase, the method uses an FCN to densely predict the pixels of the blob lines that strike through the text lines (\figurename~\ref{phases}(b)). In the second phase, we use an EM framework to extract the pixel labels of text lines with the assistance of detected blob lines (\figurename~\ref{phases}(c)). In the rest of this section we give a detailed of description FCN, EM and how they are used for text line detection and text line extraction.

\subsection{Text line detection using FCN}
Fully Convolutional Network (FCN) is an end-to-end semantic segmentation algorithm that extracts the features and learns the classifier function simultaneously. FCN inputs the original images and their pixel level annotations for learning the hypothesis function that can predict whether a pixel belongs to a text line label or not. A crucial decision have to be made about the representation of text line detection. Text line detection labels can be represented as baselines or blob lines. 

\begin{figure}[h]
\centering
\includegraphics[width=7cm]{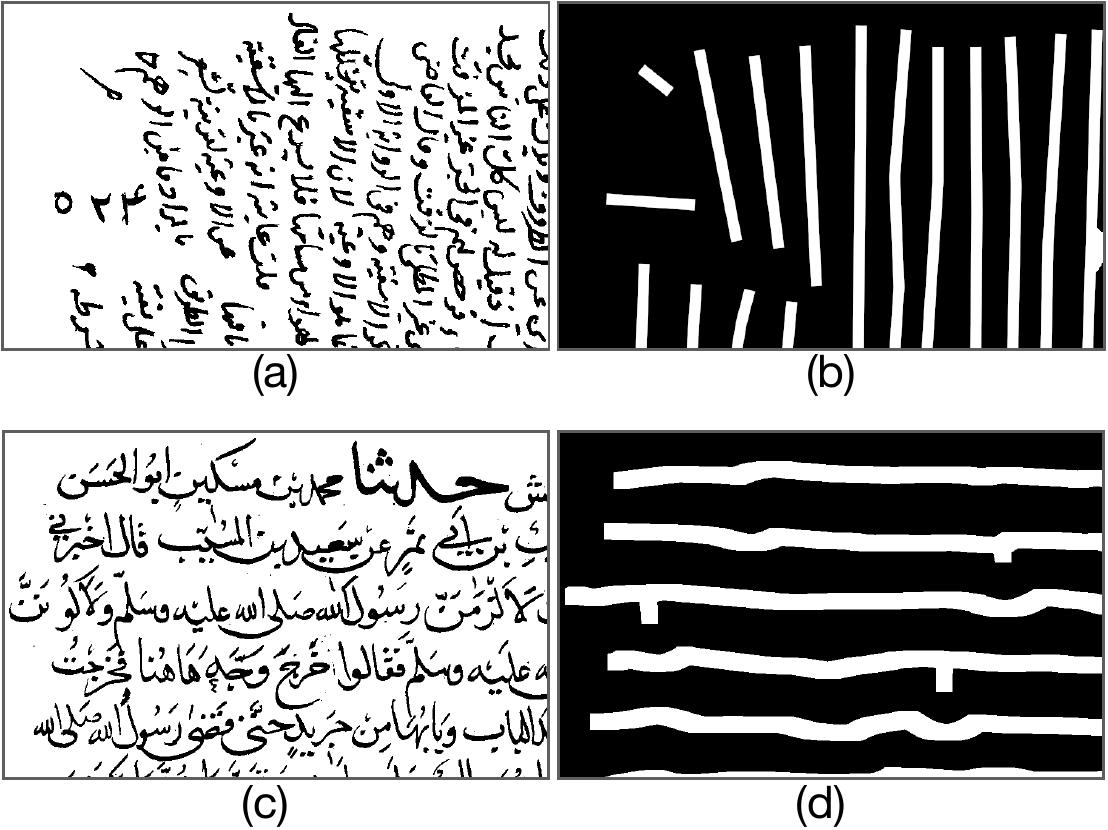}
\caption{Sample patches from document images of VML-MOC (a) and VML-AHTE (c). Blob line labeling for VML-AHTE and DIVA-HisDB is generated automatically (d). Blob line labeling for VML-AHTE is manually drawn using a paint brush with a diameter of 12 pixels (b).}
\label{manual_automatic_labels}		
\end{figure}

We use blob line labeling that connects the characters in the same line while disregarding diacritics and touching components among the text lines. Blob line labeling for VML-AHTE and DIVA-HisDB datasets is automatically generated using the skeletons of bounding polygons provided by their ground truth (\figurename~\ref{manual_automatic_labels}(d)). Blob line labeling for VML-MOC dataset is manually drawn using a sharp rectangular brush with a diameter of 12 pixels (\figurename~\ref{manual_automatic_labels}(b)).

\subsubsection{FCN architecture}
\begin{figure}[h]
\centering
\includegraphics[width=10cm]{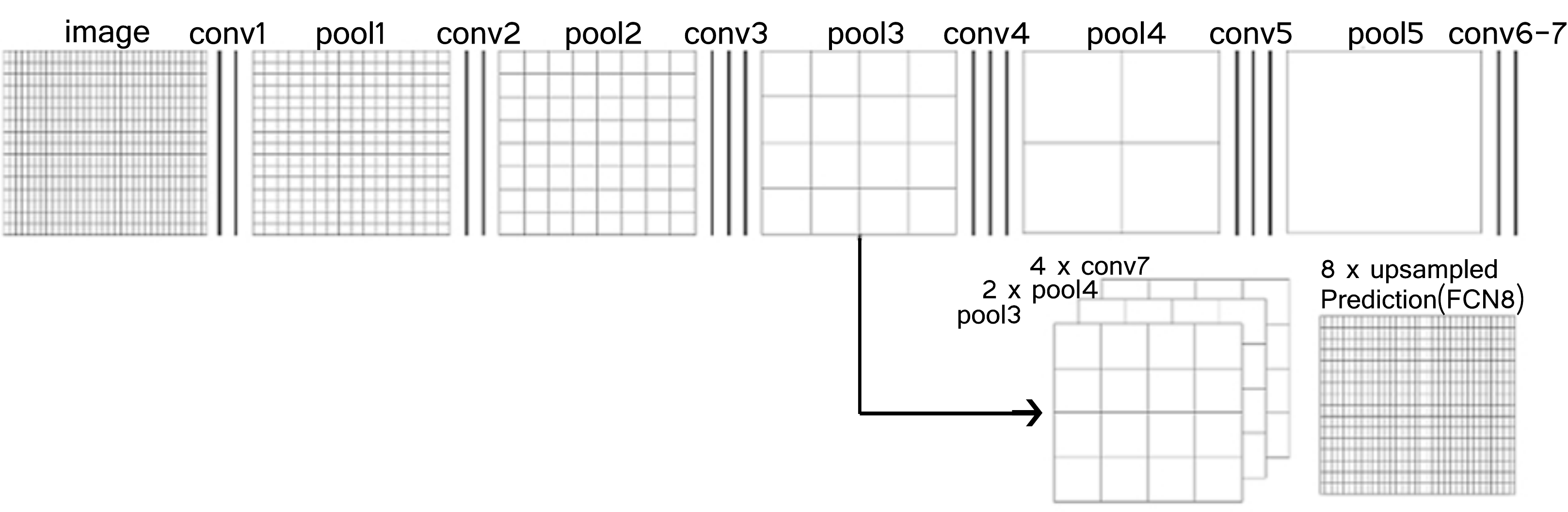}
\caption{The FCN architecture used for text line detection. Vertical lines show the convolutional layers. Grids show the relative coarseness of the pooling and prediction layers. FCN8 upsamples 4 times the final layer, upsamples twice the pool4 layer, and combine them with pool3 layer. Finally, FCN8 upsamples the combination to the input size.}
\label{arch}
\end{figure}

The FCN architecture (\figurename~\ref{arch}) we used is based on the FCN8 proposed for semantic segmentation \cite{long2015fully}. Particularly FCN8 architecture was selected because it has been successful in page layout analysis of  handwritten documents \cite{kurar2018binarization}. It consists of an encoder and a decoder. The encoder downsamples the input image and the filters can see coarser information with larger receptive field. Consequently the decoder adds final layer of encoder to the lower layers with finer information, then upsamples the combined layer back to the input size. Default input size is $224\times224$, which does not cover more than 2 to 3 text lines. To include more context, we changed the input size to $350\times350$ pixels. We also changed the number of output channels to 2, which is the number of classes: blob line or not.

\subsubsection{FCN training}
For training, we randomly crop $50,000$ patches of size $350\times350$ from inverted binary images of the documents and their corresponding labels from the blob line label images (\figurename~\ref{manual_automatic_labels}(b)). We adopted this patch size due to memory limitation. Using full pages for training and prediction is not feasible on non-specialized systems without resizing the pages to a more manageable size. Resizing the pages will result in details loss, which usually reduces the accuracy of segmentation results.

The FCN was trained by a batch size of 12, using Stochastic Gradient Descent (SGD) with momentum equals to $0.9$ and learning rate equals to $0.001$. The encoder part of FCN was initialized with its publicly available pre-trained weights.

\subsubsection{FCN testing}
During the testing, a sliding window of size $350\times350$ was used for prediction, but only the inner window of size $250\times250$ was considered to eliminate the edge effect. Page was padded with black pixels at its right and bottom sides if its size is not an integer multiple of the sliding window size, in addition to padding it at 4 sides for considering only the central part of the sliding window.

\subsection{Text line extraction using EM}
We adapt the energy minimization (EM) framework \cite{boykov2001fast} that uses graph cuts to approximate the minima of arbitrary functions. These functions can be formulated in terms of image elements such pixels or connected components. In this section we formulate a general function for text line extraction using text line detection. Then, we adapt this general function to be used with connected components for text line extraction.

\subsubsection{EM formulation}
Let $\mathcal{L}$ be the set of binary blob lines, and $\mathcal{E}$ be the set of elements in the binary document image. Energy minimization finds a labeling $f$ that assigns each element $e\in \mathcal{E}$ to a label $l_e\in \mathcal{L}$, where energy function $\textbf{E}(f)$ has the minimum. 

\begin{equation}
    \textbf{E}(f) = \sum_{e\in {\mathcal E}}D(e, \ell_e)+\sum_{\{e,e'\}\in \mathcal N}d(e, e')\cdot \delta (\ell_e \neq \ell_{e'})
\label{eq:em}
\end{equation}

The term $D$ is the data cost, $d$ is the smoothness cost, and  $\delta$ is an indicator function. Data cost is the cost of assigning element $e$ to label $l_e$. $D(e, \ell_e)$ is defined to be the Euclidean distance between the centroid of the element $e$ and the nearest neighbour pixel in blob line $l_e$ for the centroid of the element $e$. 
Smoothness cost is the cost of assigning neighbouring elements to different labels. Let $\mathcal{N}$ be the set of nearest element pairs. Then $\forall \{e,e'\}\in \mathcal {N}$,
\begin{equation}
    d(e,e') = \exp({-\beta\cdot d_e(e,e')})
\label{eq:sc}
\end{equation}
 where $d_e(e,e')$ is the Euclidean distance between the centroids of the elements $e$ and $e'$,  and $\beta$ is defined as
 \begin{equation}
     \beta=(2\left<d_e(e,e')\right>)^{-1}
 \end{equation}
  $\left<\cdot\right>$ denotes expectation over all pairs of neighbouring elements \cite{boykov2001interactive} in a document page image. $\delta (\ell_e \neq \ell_{e'})$ is equal to $1$ if the condition inside the parentheses holds and $0$ otherwise. 

\subsubsection{EM adaptation to connected components}
The presented method extracts text lines using results of the text line detection procedure by FCN. Extraction level representation labels each pixel of the text lines in a document image. The major difficulty in pixel labeling lies in the computational cost. A typical document image in the experimented datasets includes around $14,000,000$ pixels. Due to this reason, we adapt the energy function (Eq. \ref{eq:em}) to be used with connected components for extraction of text lines.

\begin{figure}[h]
\centering
\includegraphics[width=7cm]{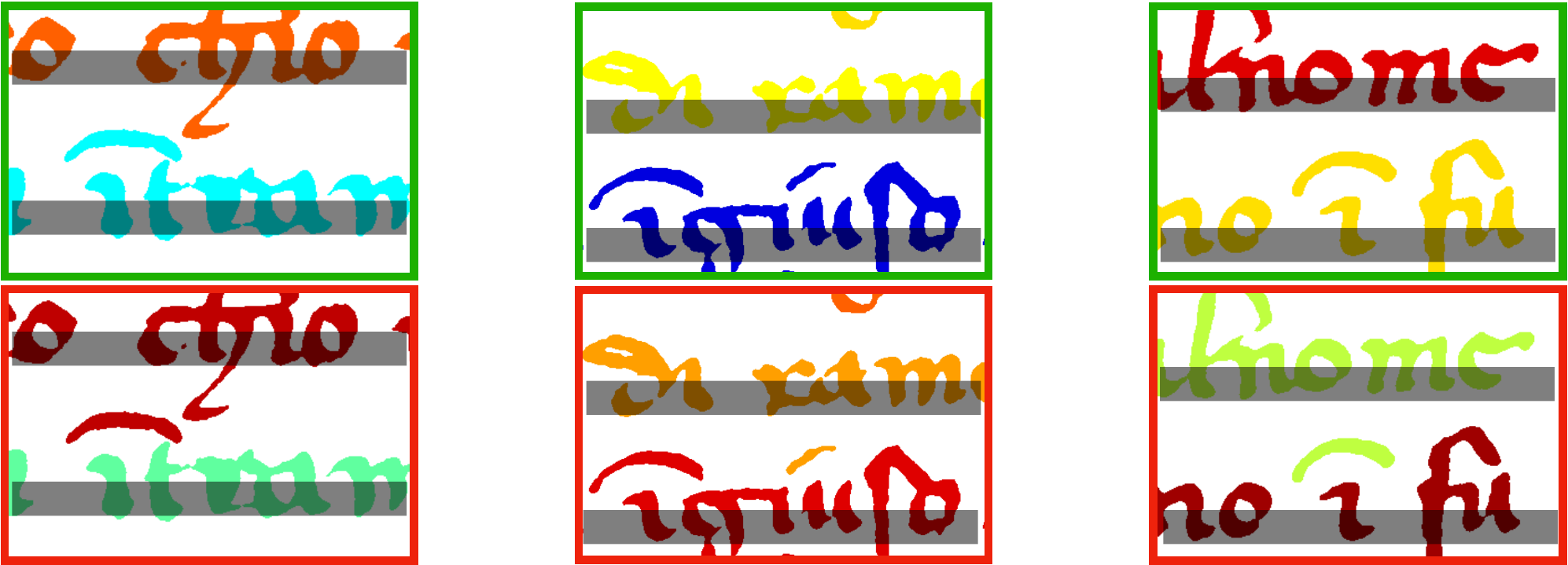}
\caption{Segmented samples that show the necessity of smoothness cost for text line extraction. Samples in the first row are true and achieved with smoothness cost. Samples in the second row are false and caused by the lack of a smoothness cost. Notice that smoothness cost pulls the diacritics to the components they belong to, in spite of their proximity to the wrong blob line.}
\label{coarse_smooth_with_blobs}
\end{figure}

Data cost of the adapted function measures how appropriate a label is for the component $e$, given the blob lines $\mathcal{L}$. Actually, the data cost alone would be equal to clustering the components with their nearest neighbour blob line. However, simply nearest neighbour clustering would be deficient to correctly label the free components that are disconnected from the blob lines (Fig. \ref{coarse_smooth_with_blobs}). 

A free component tends to exist closer to the components of a line it belongs to, but can be a nearest neighbour of a blob line that it does not belong to. This is because the proximity grouping strength decays exponentially with Euclidean distance \cite{kubovy2008whole}. This phenomenon is formulated using the smoothness cost (Eq. \ref{eq:sc}). Semantically this means that closer components have a higher probability to have the same label than distant components. Hence, the competition between data cost and smoothness cost dictates free components to be labeled spatially coherent with their neighbouring components.

\section{Experiments}
\label{experiments}

We experiment with three datasets that are different in terms of the text line segmentation challenges they contain. VML-AHTE dataset exhibits crowded diacritics and cramped text lines, whereas DIVA-HisDB dataset contains consequently touching text lines. Completely different than them VML-MOC exhibits challenges caused by arbitrarily skewed and curved text lines. The performance is measured using the line segmentation evaluation metrics of ICDAR 2013 \cite{gatos2010icfhr} and ICDAR 2017 \cite{alberti2017open}.

\subsection{ICDAR 2013 line segmentation evaluation metrics}
ICDAR 2013 metrics calculate recognition accuracy ($RA$), detection rate ($DR$) and F-measure ($FM$) values. Given a set of image points $I$, let $R_i$ be the set of points inside the $i^{th}$ result region, $G_j$ be the set of points inside the $j^{th}$ ground truth region, and $T(p)$ is a function that counts the points inside the set $p$, then the $MatchScore(i,j)$ is calculated by \equationautorefname~\ref{match}
\begin{equation}
  MatchScore(i,j) = \frac{T(G{j}\cap R{i})}{T(G{j}\cup R{i})} 
 \label{match}
\end{equation}
The evaluator considers a region pair $(i,j)$ as a one-to-one match if the  $MatchScore(i,j)$ is equal or above the threshold, which we set to $90$. Let $N_1$ and $N_2$ be the number of ground truth and output elements, respectively, and let $M$ be the number of one-to-one matches. The evaluator calculates the $DR$, $RA$ and $FM$ as follows:
\begin{equation}
  DR = \frac{M}{N_1} 
\end{equation}
\begin{equation}
  RA = \frac{M}{N_2} 
\end{equation} 
\begin{equation}
  FM=\frac{2\times DR\times RA}{DR+RA} 
\end{equation}
 
\subsection{ICDAR 2017 line segmentation evaluation metrics}
ICDAR 2017 metrics are based on the Intersection over Union (IU). IU scores for each possible pair of Ground Truth (GT) polygons and Prediction (P) polygons are computed as follows: 
\begin{equation}\label{iu}
    IU=\frac{IP}{UP}
\end{equation}
IP denotes the number of intersecting foreground pixels among the pair of polygons. UP denotes number of foreground pixels in the union of foreground pixels of the pair of polygons. The pairs with maximum IU score are selected as the matching pairs of GT polygons and P polygons. Then, pixel IU and line IU are calculated among these matching pairs. For each matching pair, line TP, line FP and line FN are given by: Line TP is the number of foreground pixels that are correctly predicted in the matching pair. Line FP is the number of foreground pixels that are falsely predicted in the matching pair. Line FN is the number of false negative foreground pixels in the matching pair.
Accordingly pixel IU is: 
\begin{equation}\label{piu}
    \text{Pixel } IU=\frac{TP}{TP+FP+FN}
\end{equation}
where TP is the global sum of line TPs, FP is the global sum of line FPs, and FN is the global sum of line FNs.

Line IU is measured at line level. For each matching pair, line precision and line recall are:
\begin{equation}\label{lineprecision}
    \text{Line precision}=\frac{\text{line } TP}{\text{line } TP + \text{line } FP}
\end{equation}
\begin{equation}\label{linerecall}
    \text{Line recall}=\frac{\text{line } TP}{\text{line } TP + \text{line } FN}
\end{equation}

Accordingly, line IU is:
\begin{equation}\label{liu}
    \text{Line } IU=\frac{\text{CL}}{\text{CL+ML+EL}}
\end{equation}
where CL is the number of correct lines, ML is the number of missed lines, and EL is the number of extra lines. 

For each matching pair: A line is correct if both, the line precision and the line recall are above the threshold value. A line is missed if the line recall is below the threshold value. A line is extra if the line precision is below the threshold value.
\subsection{Results on VML-AHTE dataset}
Since VML-AHTE and VML-MOC datasets are recently published datasets we run two other supervised methods. First method is a holistic method that can extract text lines in one phase and is based on instance segmentation using MRCNN \cite{he2017mask}. Second method is based on running the EM framework using the blob line labels from the ground truth and we refer to it Human+EM. On VML-AHTE dataset, FCN+EM outperforms all the other methods in terms of all the metrics except Line IU. It can successfully split the touching text lines and assign the disjoint strokes to the correct text lines.
\begin{table}[h]
    \centering
    {
    \centering
    \caption{Results on VML-AHTE dataset}
    \begin{tabular}{l c | c | c | c | c}
        \toprule
        \multicolumn{1}{l}{} & Line IU & Pixel IU & DR & RA & FM\\
        \hline
        FCN+EM & 94.52 & \textbf{90.01} & \textbf{95.55} & \textbf{92.8} & \textbf{94.3}\\
        MRCNN & 93.08 & 86.97 & 84.43 & 58.89 & 68.77\\
        Human+EM & \textbf{97.83} & 89.61 & 88.14 & 87.78 & 87.96\\
        \bottomrule
    \end{tabular}
    \label{ahte_results}
    }
\end{table}

\begin{figure}
    \centering
    \includegraphics[width=7cm]{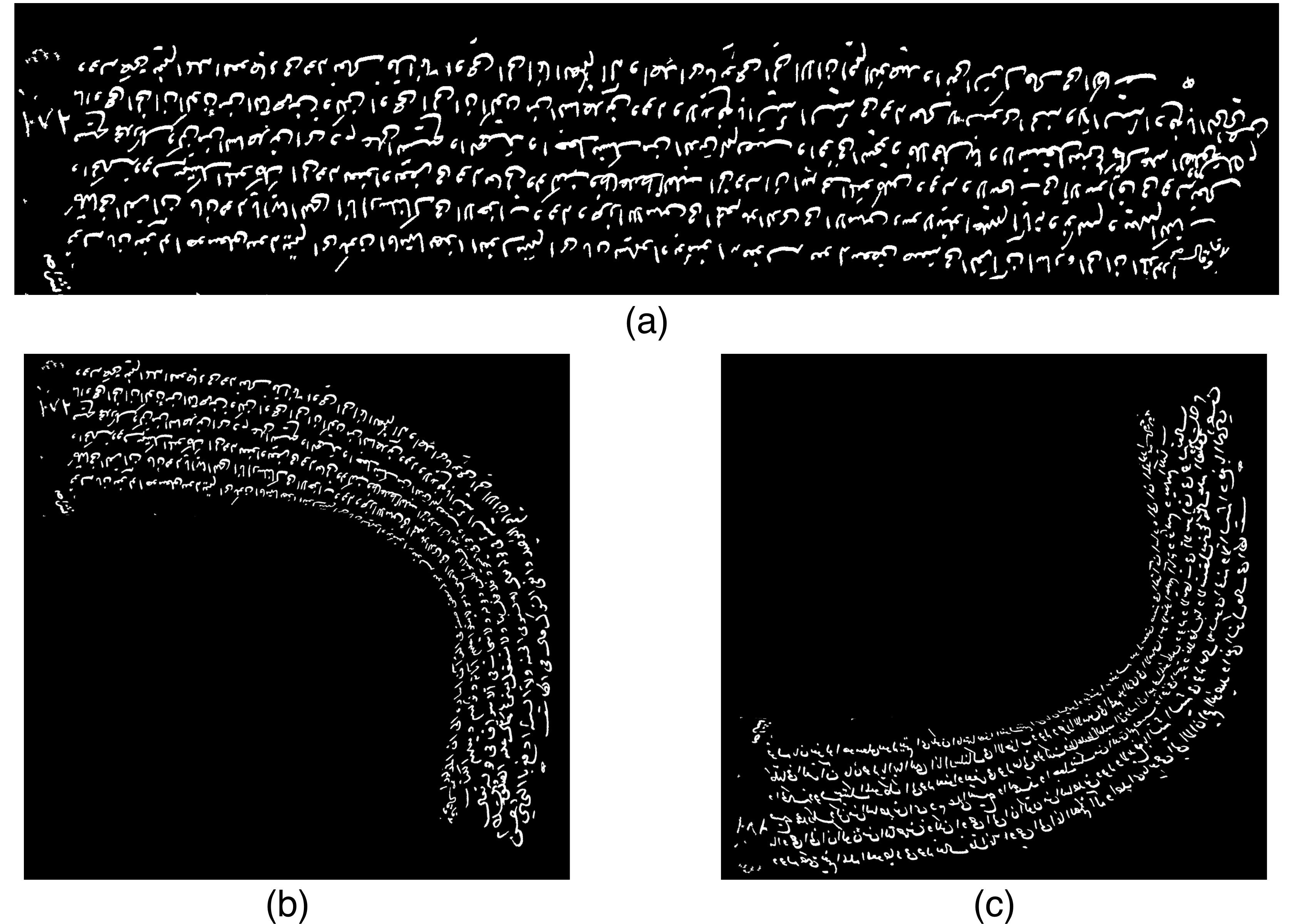}
    \caption{Example of generated curved lines: (a) shows the original straight lines section, (b) is the result of warping (a) by 90 degrees in the middle to generated the curved lines, and (c) is the mirrored image of (b) in the vertical direction.}
    \label{fig:augmentation}
\end{figure}

\subsection{Results on VML-MOC dataset}
The VML-MOC dataset contains both types, straight text lines and curved text lines. Number of straight text lines is substantially greater than the number of curved text lines. This imbalance causes the FCN to overfit on the straight text lines. This in turn leads to fragmented blob lines when predicting over the curved text lines. Therefore, to compensate this imbalance, we generated images containing artificially curved text lines. We selected the document image parts with straight lines and warp these images $90$ degrees from their middle. Furthermore, each one of those warped lines was mirrored in the horizontal and vertical directions resulting in curved lines in four directions. Figure \ref{fig:augmentation} illustrates this procedure. The FCN+EM that is trained with augmented curved text lines (FCN+EM+Aug) outperforms the FCN+EM that is trained only with the training set (Table~\ref{moc_results}). But FCN+EM+Aug still underperforms a learning free algorithm \cite{barakat2019vml}.
\begin{table}[h]
    \centering
    {
    \centering
    \caption{Results on VML-MOC dataset}
    \begin{tabular}{l c | c | c | c | c}
        \toprule
        \multicolumn{1}{l}{} & Line IU & Pixel IU & DR & RA & FM\\
        \hline
        FCN+EM          & 25.04  & 48.71 & 26.45 & 17.73 & 31.09\\
        FCN+EM+Aug      & 35.12  & 60.97 & 84.43 & 58.89 & 68.77\\
        \cite{barakat2019vml} & 60.99  & 80.96 & -     & -     & -\\
        Human+EM      & \textbf{96.62}  & \textbf{99.01 } &\textbf{90.41}  &\textbf{91.74}  &\textbf{91.03} \\
        \bottomrule
    \end{tabular}
    \label{moc_results}
    }
\end{table}

\subsection{Results on DIVA-HisDB dataset}
We compare the results with the results of Task-3 from ICDAR 2017 competition on layout analysis for medieval manuscripts~\cite{simistira2017icdar2017}. Task-3's scope of interest is only the main text lines but not the interlinear glosses. We removed these glosses prior to all our experiments using the ground truth. It should be noticed that Task-3 participants removed these glosses using their own algorithms.

Table~\ref{diva_results} presents a comparison of our methods with the participants of ICDAR 2017 competition on layout analysis for challenging medieval manuscripts for text line extraction. 
The FCN+EM can obtain a perfect Line IU score on the books CSG863 and CB55. Its Pixel IU is on par with the best preforming method in the competition.
\begin{table}[h]
\centering
    {\centering
    \caption{Comparison with the Task-3 results of the ICDAR2017 competition on layout analysis for challenging medieval manuscripts~\cite{simistira2017icdar2017}.}
    \begin{tabular}{l c c | c c | c c}
        \toprule
        \multicolumn{1}{l}{} &\multicolumn{2}{c|}{CB55} &  \multicolumn{2}{c|}{CSG18} & \multicolumn{2}{c}{CSG863}\\
        \hline
        \multicolumn{1}{c}{} & LIU & PIU & LIU & PIU & LIU & PIU\\
        \hline
        FCN+EM & \textbf{100} & \textbf{97.64} & \textbf{97.65} & \textbf{97.79} & \textbf{100} & 97.18\\
        System-2 & 84.29 & 80.23 & 69.57 & 75.31 & 90.64 & 93.68\\
        System-6  & 5.67 & 30.53 & 39.17 & 54.52 & 25.96 & 46.09\\
        System-8  & 99.33 & 93.75 & 94.90 & 94.47 & 96.75 & 90.81\\
        System-9+4.1  & 98.04 & 96.67 & 96.91 & 96.93 & 98.62 & \textbf{97.54}\\
        \bottomrule
    \end{tabular}
    \label{diva_results}
    }
\end{table}

\subsection{Discussion}
\begin{figure}[h]
\centering
\includegraphics[width=8cm]{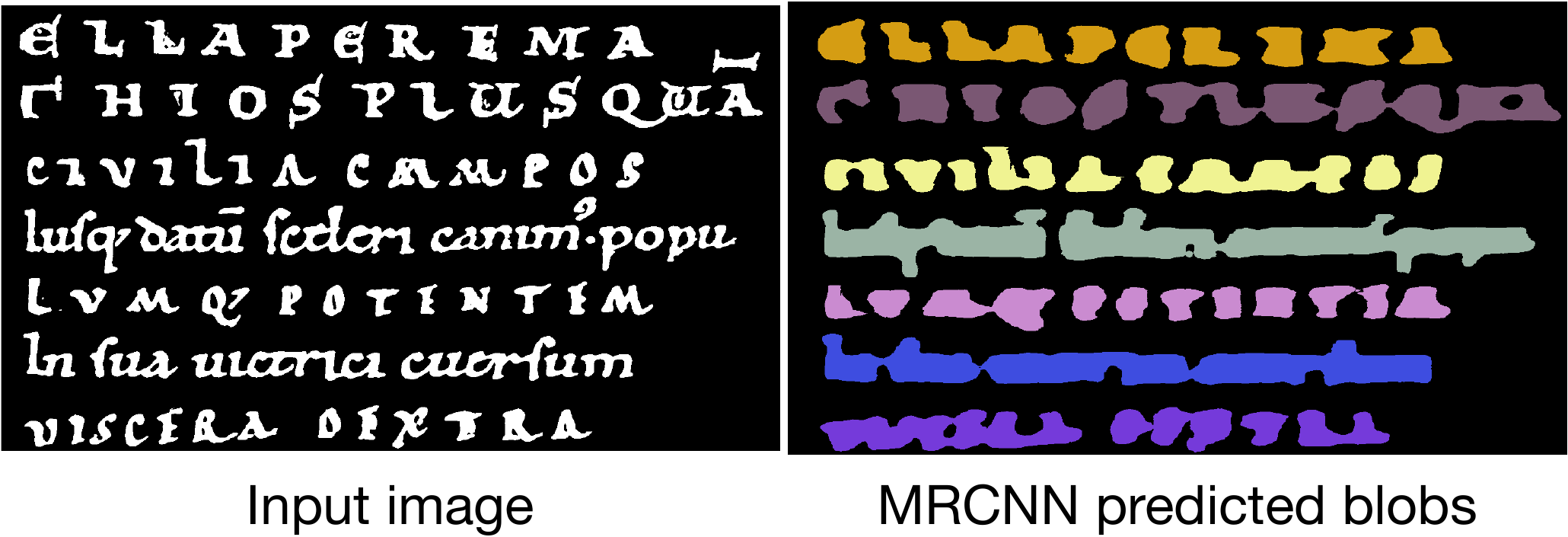}
\caption{MRCNN method correctly predicts text line pixels but its results are not fairly evaluated due to disconnected polygons.}
\label{mrcnn_seperate_blobs}		
\end{figure}

An observable pattern in the results is the parallel flow of line IU values and pixel IU values while RA values are fluctuating in comparison to DR values. Clearly, such counter-intuitive behaviour of a metric is not preferable in terms of interpretability of the results. On the other hand, ICDAR 2017 evaluator can not handle the cases where a text line consists of multiple polygons. Such case arises from MRCNN results. MRCNN segments a text line instance correctly but represents it as multiple polygons with the same label. Evaluating MRCNN results in their raw form yields to low values unfairly (Figure~\ref{mrcnn_seperate_blobs}). Because ICDAR 2017 evaluator calculates an IU score for each possible pair of ground truth polygons and prediction polygons then selects the pairs with maximum IU score as the matching pairs. Consequently, a text line represented by multiple polygons is considered only by its largest polygon. 

\section{Conclusion}
\label{conclusion}
This paper presents a supervised text line segmentation method FCN+EM. The FCN detect the blob lines that strike through the text lines and the EM extracts the pixels of text lines with the guidance of the detected blob lines. FCN+EM does not make any assumption about the text line orientation or text line height. The algorithm is very effective in detecting cramped, crowded and touching text lines. It has a superior performance on VML-AHTE and DIVA-HisDB datasets but comparable results on VML-MOC dataset.

\section*{Acknowledgment}
The authors would like to thank Gunes Cevik for annotating the ground truth. This work has been partially supported by the Frankel Center for Computer Science.

\bibliographystyle{splncs04}
\bibliography{xref.bib}

%
%
%

\end{document}